\documentclass{article}

\usepackage{microtype}
\usepackage{graphicx}
\usepackage{multirow}
\usepackage{subfigure}
\usepackage{stfloats}
\usepackage{booktabs}
\usepackage{hyperref}

\usepackage{amsmath}
\usepackage[accepted]{icml2020}

\icmltitlerunning{Unsupervised Transfer Learning for Spatiotemporal Predictive Networks}

\newcommand{\eqn}[1]{Eq.~(\ref{#1})}

\newcommand{\tab}[1]{Table~\ref{#1}}
\newcommand{\fig}[1]{Figure~\ref{#1}}

\begin{document}

\twocolumn[
\icmltitle{Unsupervised Transfer Learning for Spatiotemporal Predictive Networks}

\icmlsetsymbol{equal}{*}

\begin{icmlauthorlist}
\icmlauthor{Zhiyu Yao}{equal,to}
\icmlauthor{Yunbo Wang}{equal,to}
\icmlauthor{Mingsheng Long}{to}
\icmlauthor{Jianmin Wang}{to}
\end{icmlauthorlist}

\icmlaffiliation{to}{School of Software, BNRist, Research Center for Big Data, Tsinghua University}

\icmlcorrespondingauthor{Mingsheng Long}{mingsheng@tsinghua.edu.cn}

\icmlkeywords{Transfer Learning, Unsupervised Learning, Spatiotemporal Prediction}

\vskip 0.3in
]

\printAffiliationsAndNotice{\icmlEqualContribution}

\begin{abstract}

This paper explores a new research problem of unsupervised transfer learning across multiple spatiotemporal prediction tasks. Unlike most existing transfer learning methods that focus on fixing the discrepancy between supervised tasks, we study how to transfer knowledge from a zoo of unsupervisedly learned models towards another predictive network. Our motivation is that models from different sources are expected to understand the complex spatiotemporal dynamics from different perspectives, thereby effectively supplementing the new task, even if the task has sufficient training samples. Technically, we propose a differentiable framework named \textit{transferable memory}. It adaptively distills knowledge from a bank of memory states of multiple pretrained RNNs, and applies it to the target network via a novel recurrent structure called the \textit{Transferable Memory Unit} (TMU). Compared with finetuning, our approach yields significant improvements on three benchmarks for spatiotemporal prediction, and benefits the target task even from less relevant pretext ones.

\end{abstract}

\section{Introduction}

Existing transfer learning methods mainly focus on how to fix the discrepancy between supervised tasks. However, unsupervised learning has achieved remarkable advances in recent years and has become a hot topic in the deep learning community. Then new questions arise: \emph{Is it necessary to do transfer learning between unsupervised tasks, and how to do it?}

As a typical unsupervised learning paradigm, predictive learning has shown great research significance in discovering the underlying structure of unlabeled spatiotemporal data without human supervision and learning generalizable deep representations from the consequences of complex video events. The studies of the spatiotemporal predictive learning can benefit many practical applications and downstream tasks, such as precipitation nowcasting \cite{shi2015convolutional}, traffic flow prediction \cite{xu2018predcnn}, physical scene understanding \cite{wu2017learning}, early activity recognition \cite{wang2019eidetic}, deep reinforcement learning \cite{ha2018recurrent}, and vision-based model predictive control \cite{finn2017deep}. Different from all the above work, in this paper, we explore how to transfer knowledge from a zoo of pretrained models towards a novel predictive learning task. Models from both the source and target domains are trained to predict sequences of future frames.

Transferring knowledge across these tasks is yet to be explored, but important. In many scenarios, deep networks may suffer from the serious problem of long-tail data distribution in the target domain. A natural solution is to finetune another model that was well-pretrained with large-scale and more effective training data. For example, when we train precipitation forecasting models for arid areas, we may exploit the laws of weather changes that are learned from other areas with abundant rainfall.
However, it is a challenging transfer learning problem, because, in the first place,  not all knowledge of the pretrained models can be directly applied to the target task due to the discrepancy of various domains. 
After all, different areas may have their unique climate characteristics.
We have to explore how to distill the transferable representations from the pretrained models without labeled data.
In the second place, with a zoo of source models, we need to dynamically adjust their impact on the training process of the target network.

To solve these problems, we propose a novel differentiable framework named \textit{transferable memory} along with the new \textit{Transferable Memory Unit} (TMU). 
Different from finetuning, our approach enables the target model to adaptively learn from a zoo of source models. It provides diverse understandings of the underlying, complex data structure of the target domain.
Technically, we perform unsupervised knowledge distillation on the memory states of multiple pretrained recurrent networks, and then introduce a new gating mechanism to dynamically find the transferable part of the distilled representations. 
In this way, we use the spatiotemporal dynamics of source domains as the prior knowledge of the final predictive model, so it can focus more on the domain-specific data structure on the target dataset.
The \textit{transferable memory} framework significantly outperforms previous transfer learning methods on three benchmarks with nine sub-datasets: a synthetic flying digits benchmark, a real-world human motion benchmark, and a precipitation nowcasting benchmark.

The contributions of this paper are summarized as follows:
\begin{itemize}
  \label{question}
    \item We introduce a new research problem of unsupervised transfer learning across multiple spatiotemporal prediction tasks. It is challenging as the data distribution of different domains can be distant, \textit{e.g.}, from various data sources, or from synthetic data to real data. 
    \item We propose a deep learning solution which features the \textit{transferable memory} and is shown effective for a wide range of RNNs, including ConvLSTM \cite{shi2015convolutional}, PredRNN \cite{wang2017predrnn}, MIM \cite{WangZZLWY19}, and SAVP \cite{lee2018stochastic}, covering both deterministic and stochastic models.
    \item We validate the effectiveness of the proposed approach on three benchmarks with a variety of data sources and have a series of empirical findings. 
    Unlike supervised transfer learning where irrelevant source data may lead to negative transfer learning effects, the proposed approach can adaptively transfer temporal dynamics from source videos even if the content seems less relevant.
\end{itemize}

\section{Related Work} 

\subsection{Spatiotemporal Prediction}

Due to the modeling capability of temporal dependencies, the early literature suggested using RNN-based models for spatiotemporal predictive learning \cite{Ranzato2014Video,srivastava2015unsupervised,Oh2015Action,de2016dynamic}.
\citet{shi2015convolutional} proposed the convolutional LSTM (ConvLSTM) that combines the advantages of the convolutions and the LSTMs to capture the spatial and temporal correlations simultaneously.
\citet{wang2017predrnn} introduced the ST-LSTM that allows the memory state to be updated across the stacked recurrent layers along a zigzag state transition path.
\citet{villegas2017learning} proposed a framework for long-term video generation with a combination of LSTMs and a pose estimation model.
\citet{wichers2018hierarchical} extended the work from \citet{villegas2017learning} by learning hierarchical video representations in an unsupervised manner.
\citet{Finn2016Unsupervised} presented a recurrent model based on ConvLSTM to predict how the content of the pixels moves instead of estimating the variations of the pixel values.
\citet{wang2019eidetic} introduced the E3D-LSTM that combines ST-LSTM, the 3D convolution, and a memory attentive module. It builds a memory-augmented recurrent network that can capture long-term video dynamics.
\citet{WangZZLWY19} treated the predictive learning task as a spatiotemporal non-stationary process and proposed to reduce the non-stationarity by replacing the forget gate of ST-LSTM with an inner recurrent structure. 
There are many other methods focusing on improving the RNN-based predictive models for spatiotemporal data \cite{Kalchbrenner2017Video,liu2017video,Villegas2017Decomposing}.
Besides these deterministic video prediction models, some recent literature explored the video prediction problem by modeling the future uncertainty. These models are either based on adversarial training \cite{Mathieu2015Deep,vondrick2016generating,tulyakov2018mocogan} or variational autoencoders (VAEs) \cite{babaeizadeh2017stochastic,tulyakov2018mocogan,denton2018stochastic}, or both \cite{lee2018stochastic,villegas2019high}.

Note that most of the above models, including both stochastic and non-stochastic models, are based on recurrent architectures such as LSTMs. Thus, in this paper, we focus on finding a transfer learning approach particularly designed for LSTM-based predictive networks, while most existing transfer learning techniques are designed for CNNs.

\subsection{Transfer Learning}

Transfer learning focuses on storing knowledge while solving one problem and applying it to a different but related problem \cite{long2015learning}. 
The ImageNet \cite{DengDSLL009} pretrained CNNs have greatly benefited many computer vision tasks such as image classification, object detection, and segmentation. 
\citet{donahue2013decaf} proposed a method to leverage the pretrained models, which directly trains a classifier upon the fixed, pretrained CNNs on the target dataset.
Apart from the initialization with the
pretrained model, \citet{LiGD18,Delta} presented several regularization techniques to retain the features learned on the source task, explicitly enhancing the similarity of the final model and the initial one. 
\citet{Rebuffi17,rebuffi-cvpr2018} introduced convolutional adapter modules upon pretrained ResNet \cite{HeZRS16} or VGGNet \cite{SimonyanZ14a} that can adapt the domain-specific knowledge from novel tasks.
\citet{LiuPS19} developed a model transfer framework named knowledge flow, in which the knowledge is transferred by intermediate features flowing from multiple pretrained teacher CNNs to a randomly initialized student CNN. To make the student CNN independent, it uses a curriculum learning strategy and gently increases the weights of features by the student compared to those by the teachers.

This work is also inspired by the idea of knowledge distillation \cite{li2014learning,hinton2015distilling}, which transfers knowledge from larger models into smaller, faster models without losing too much generalization ability.
\citet{romero2015fitnets} and \citet{ zagoruyko2017paying} proposed to explicitly produce similar response patterns in the teacher and student feature maps. 
The above work and many other papers \cite{huang2017like,yim2017gift,kim2018paraphrasing,koratana2019lit,ahn2019variational} mainly focus on distilling knowledge to solve the model compression problem within the same dataset. These methods were not designed for cross-domain transfer learning and are therefore different from our approach.
\citet{gupta2016cross} introduced a cross-modal knowledge distillation technique to transfer supervision between images from different modalities, while our transfer learning approach is unsupervised.

In contrast with all the above transfer learning methods designed for CNNs, we focus on the transfer learning problem for predictive RNNs. This problem is under-explored, especially in spatiotemporal scenarios.
In the field of natural language processing, \citet{CuiZSJW19} proposed a recurrent transfer learning framework that transfers hidden states from the teacher RNN to the student RNN. However, upon training, this method still relies on the pretrained teacher models, and thus requires extra memory footprint. 
Different from this work, our paper presents a novel framework for a new problem, \textit{i.e.}, transferring knowledge across multiple unsupervised prediction tasks for spatiotemporal data.

\begin{figure*}[t]
    \vspace{5pt}
    \centering
    \includegraphics[width=0.66\textwidth]{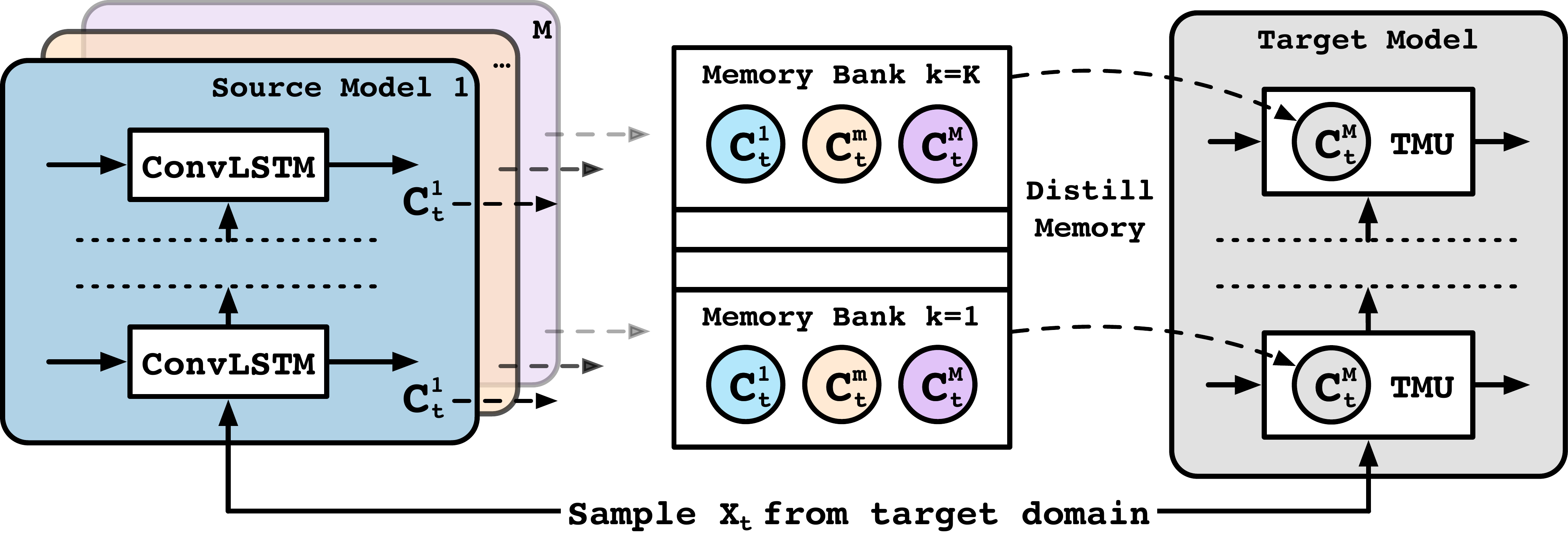}
    \vspace{-5pt}
    \caption{An overview of the \textit{transferable memory} framework, which learns a predictive network on the target dataset from $M$ pretrained networks that were collected from different sources. $K$ is the number of the recurrent layers. Without loss of generality, we use the ConvLSTM \cite{shi2015convolutional} for the source models, yet this framework can be applied to any variants of LSTMs.}
    \label{framework}
\end{figure*}

\section{Method}

In this section, we provide a solution to the problem of distilling knowledge from unsupervisedly pretrained predictive networks, and transferring it to a new spatiotemporal prediction task. Different from most previous work in transfer learning, our approach is specifically designed for RNN models and unlabelled sequential data. Below we introduce the overall \textit{transferable memory} framework, a new recurrent unit named TMU, and the multi-task training objective for knowledge distillation and sequence prediction.

\subsection{Transferable Memory Framework}

\paragraph{Why transfer memory representations?}

The memory state of the LSTM unit \cite{Hochreiter1997Long} can latch the gradients during the training process of the recurrent networks, to alleviate the gradient vanishing problem, thereby storing valuable information about the underlying temporal dynamics. In spatiotemporal predictive learning scenarios, the effectiveness of the memory states has also been explored and validated \cite{wang2017predrnn}. They are important for multi-step future prediction as they convey long-term features of the spatiotemporal data.
Besides, training an LSTM-based model in the predictive learning manner, \textit{i.e.}, one of the unsupervised learning paradigms, has been empirically proved to successfully learn concept-level representations that can benefit downstream supervised tasks \cite{wang2019eidetic}. Therefore, we assume that the predictive networks that were pretrained on different unlabeled datasets can provide knowledge of their source domains, and understand the spatiotemporal dynamics of a new task from different perspectives. 
Now, the question is how to effectively leverage the memory representations of multiple pretrained models. \fig{framework} shows our proposed \textit{transferable memory} framework, which enables the student recurrent network to learn from $M$ existing teacher models.

\paragraph{Memory bank.}
Without loss of generality, we use ConvLSTM \cite{shi2015convolutional} as the building block of the source models. Note that the proposed framework can be easily applied to other forms of future frames prediction models, such as the Spatiotemporal LSTM \cite{wang2017predrnn}, the Video Pixel Network \cite{Kalchbrenner2017Video}, the Eidetic LSTM \cite{wang2019eidetic}, etc. In this paper, we do not focus on discussing how to pretrain the source models. 
During the training process of the target network on a new dataset, the parameters of the source models are frozen, and they are not taken as the initialization of the target model. In other words, the target model is trained from scratch. It gradually obtains knowledge from the pretrained networks via knowledge distillation. Both the source models and the target one take the same input sequences. Formally, at time step $t$, each unit of the source model computes
\begin{equation}
\label{Source-ConvLSTM}
    \mathcal{H}_t^m, \mathcal{C}_t^m = \textrm{ConvLSTM}\left(\mathcal{X}_t, \mathcal{H}_{t-1}^m, \mathcal{C}_{t-1}^m\right),
\end{equation}
where $\mathcal{X}_t$ is the input state that can be an input frame or the hidden state from the lower layer. $\mathcal{H}_t^m$, $\mathcal{C}_t^m$ are respectively the hidden state and memory state of the $m$-th pretrained networks, where $m \in\left\{{1}, \ldots, {M}\right\}$. Then we obtain the \textit{memory bank} in forms of $\left\{\mathcal{C}_{t}^1, \ldots, \mathcal{C}_{t}^M\right\}$, which contains diverse representations of the spatiotemporal dynamics, part of which can contribute to the target task.

\begin{figure*}[t]
    \vspace{5pt}
    \centering
    \includegraphics[width=0.7\textwidth]{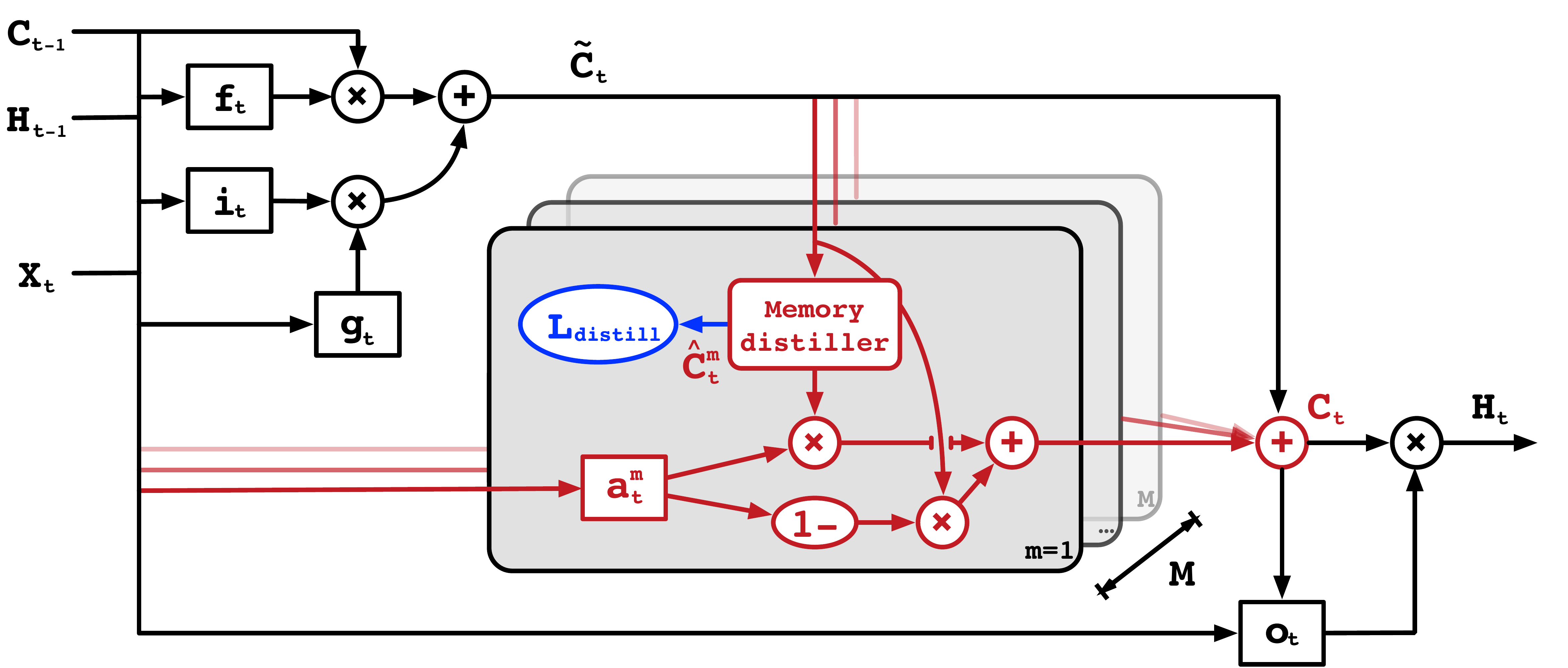}
    \caption{The architecture of the TMU that is used in the target predictive network. It unsupervisedly distills knowledge (in terms of diverse representations of the spatiotemporal dynamics) from a bank of memory states of a zoo of pretrained models. Here, $M$ is the number of pretrained models, and $a_t^m$ is the transfer gate that corresponds to the $m$-th pretrained model.}
    \label{unit}
\end{figure*}

\subsection{Transferable Memory Unit}

The \textit{Transferable Memory Unit} (TMU) is the basic building block of the target network (see \fig{framework}). It is designed to distill transferable features from the memory bank and dynamically adjust the influence of all source networks. As shown in \fig{unit}, the main architecture of TMU has three components: a \textit{memory distiller} module, a set of \textit{transfer gate}, and the basic operations following the ConvLSTM \cite{shi2015convolutional}, which are specified as follows:
\begin{equation}
\label{convlstm}
\begin{split}
    & g_{t} =\tanh \left(W_{x g} \ast \mathcal{X}_{t}+W_{h g} * \mathcal{H}_{t-1}+b_{g}\right) \\ 
    & i_{t} =\sigma\left(W_{x i} * \mathcal{X}_{t}+W_{h i} * \mathcal{H}_{t-1}+W_{c i} \odot \mathcal{C}_{t-1}+b_s\right) \\ 
    & f_{t} =\sigma\left(W_{x f} * \mathcal{X}_{t}+W_{h f} * \mathcal{H}_{t-1}+W_{c f} \odot \mathcal{C}_{t-1}+b_{f}\right) \\ 
    & \widetilde{\mathcal{C}}_t =f_{t} \odot \mathcal{C}_{t-1}+i_{t} \odot g_{t}, \\
\end{split}
\end{equation}
where $\sigma$ is sigmoid activation function, $*$ and $\odot$ denote the convolution operator and the Hadamard product respectively. Unless otherwise mentioned, all through this text, the convolutional filters are $5\times5$.
The use of the input gate $i_{t}$, forget gate $f_{t}$, and input-modulation gate $g_{t}$ controls the information flow towards the intermediate memory state $\widetilde{\mathcal{C}}_t$. Here we build TMU upon ConvLSTM for the sake of convenience, yet the proposed memory distiller and transfer gate, being displayed as a whole by the gray box in \fig{unit}, can be seamlessly integrated into any forms of LSTM-like recurrent units.

\paragraph{Memory distiller module.}

The memory distiller module is largely inspired by recent advances on \textit{compressing many visual domains in relatively small networks, with substantial parameter sharing between them} \cite{Rebuffi17,rebuffi-cvpr2018}, which have also been shown to mitigate the forgetting problem of finetuning. However, different from these existing methods that were particularly designed for transferring knowledge from a single source model to multiple target models, our memory distiller is used for vice-versa.
As shown in \fig{unit}, TMU contains $M$ memory distiller modules, corresponding to the number of source models. Each memory distiller takes $\widetilde{\mathcal{C}}_t$ as input, and employs a $1 \times 1$ convolutional layer parametrized as $W_\text{distill}^m$ for each pretext task, followed by layer normalization \cite{ba2016layer}:
\begin{equation}
\label{distiller}
    \widehat{\mathcal{C}}_t^m = \textrm{LayerNorm}\left(W_\text{distill}^m * \widetilde{\mathcal{C}}_{t}\right).
\end{equation}

We then use the generated features $\{\widehat{\mathcal{C}}_t^1, \ldots ,\widehat{\mathcal{C}}_t^M\}$ to distill knowledge from the memory bank mentioned above $\{\mathcal{C}_t^1, \ldots ,\mathcal{C}_t^M\}$. 
Over all pretext tasks and across the time horizon, we minimize the Euclidean distance between pairs of memory states:
\begin{equation}
\label{distill}
    \mathcal{L}_\text{distill} = \sum_{s=1}^{M}\sum_{t=1}^{T} \|  \widehat{\mathcal{C}}_t^m - \mathcal{C}_t^m \|_{2}^{2}.
\end{equation}

The distillation loss enables TMU to learn separately from multiple teachers, thereby gaining substantial prior knowledge of the complex spatiotemporal dynamics. 
In this way, throughout the training process, the student network can focus on more domain-specific patterns of the target dataset.
Noticeably, the target memory $\widehat{\mathcal{C}}_t^m$ would not converge to the mean of source memories as in \eqn{distiller} we have $M$ sets of parameters $W_\text{distill}^m$ to match each $\widehat{\mathcal{C}}_t^m$ with the corresponding source memory.

\paragraph{Transfer gate.}
However, two problems remain for learning from multiple pretext domains. First, not all memory representations by the source models are transferable and yield a positive effect to the target task. Second, the source models should not equally contribute to the target one. To further solve these problems, we propose to learn a set of transfer gates $\{a_t^1, \ldots ,a_t^M\}$ to adaptively control the information flow from the previously distilled memory representations $\{\widehat{\mathcal{C}}_t^1, \ldots ,\widehat{\mathcal{C}}_t^M\}$ to the final memory state ${\mathcal{C}}_t$. 
Finally, TMU obtains the output hidden state as follows. 
\begin{equation}
\label{gate}
\begin{split}
    & a_t^m =\sigma \left(W_{x m} * \mathcal{X}_{t}+W_{h m} * \mathcal{H}_{t-1} + b_m\right)\\
    & \mathcal{C}_t = \widetilde{\mathcal{C}}_t + \sum_{m=1}^M \left(a_t^m \odot \widehat{\mathcal{C}}_t^m+\left(1-a_t^m\right) \odot \widetilde{\mathcal{C}}_t \right)\\
    & o_t =\sigma\left(W_{x o} * \mathcal{X}_t+W_{h o} * \mathcal{H}_{t-1}+W_{c o} \odot \mathcal{C}_t + b_o\right) \\
    & \mathcal{H}_t = o_t \odot \tanh \left(\mathcal{C}_t\right).
\end{split}
\end{equation}

The complete computation of TMU consists of \eqn{convlstm}, \eqn{distiller}, and \eqn{gate}. When $a_t^m$ approaches $1$, more pretext knowledge is distilled from the $m$-th source domain to the learned model. By controlling the states in $\{a_t^1,\ldots,a_t^M\}$, TMU can dynamically adjust the influence of $M$ sources.

\subsection{Unsupervised Training Objective}

All the training procedures of the transfer memory framework are unsupervised. The final training objective is:
\begin{equation}
\label{final_loss}
 \mathcal{L}_\text{final} = \sum_{t=2}^{T} \|  \widehat{\mathcal{X}}_t - \mathcal{X}_t \|_{2}^{2} + \beta \sum_{k=1}^{K} \mathcal{L}_\text{distill}^k,
\end{equation}
where $\widehat{\mathcal{X}}_t$ is the generated frame, $k \in \{1,\ldots,K\}$ is the index of the TMU layer, $\mathcal{L}_\text{distill}^k$ is defined in \eqn{distill}, and $\beta$ is a hyper-parameter tuned on the target validation set. 
It is worth noting that we do not use any parameters to the distillation loss terms of different source domains in \eqn{distill}. It is because that due to \eqn{distiller}, $\{\widehat{\mathcal{C}}_t^1, \ldots, \widehat{\mathcal{C}}_t^M\}$ can learn domain-specific patterns so that $\widetilde{\mathcal{C}}_t$ can focus on common ones. Further, in \eqn{gate}, the transfer gates $\{a_t^1, \ldots, a_t^M\}$ dynamically adjust the significance of all source domains.

\section{Experiments}

We study unsupervised transfer learning performed between different spatiotemporal prediction tasks, within or across the following three benchmarks:

\paragraph{Flying digits.} This synthetic benchmark has three Moving MNIST datasets with respectively $1$, $2$, or $3$ flying digits randomly sampled from the static MNIST dataset. Each dataset contains $10{,}000$ training sequences, $2{,}000$ validation sequences, and $3{,}000$ testing sequences. Each sequence consists of $20$ consecutive frames, $10$ for the input, and $10$ for the prediction. Each frame is of the resolution of $64\times64$.

\paragraph{Human motion.} This benchmark is built upon three human action datasets with real-world videos: Human3.6M \cite{ionescu2013human3}, KTH \cite{Sch2004Recognizing}, and Weizmann \cite{BlankGSIB05}. Specifically, we use the Human3.6M dataset as the target domain, which has $2{,}220$ sequences for training, $300$ for validation, and $1{,}056$ for testing. We follow \cite{WangZZLWY19} to resize each RGB frame to the resolution of $128\times128$, and make the model predict $4$ future frames based on $4$ previous ones.

\paragraph{Precipitation nowcasting.} Precipitation nowcasting is a meaningful application of spatiotemporal prediction. This benchmark consists of three radar echo datasets\footnote{Predicting the shapes and trajectories of future radar echoes is the foundation of accurate precipitation nowcasting \cite{shi2015convolutional,shi2017deep,wang2017predrnn}.}: two of them are from separate years of Guangzhou, and the other one is from Beijing, which is a more arid place. The Guangzhou2016 dataset has $33{,}769$ consecutive radar observations, collected every $6$ minutes. The Guangzhou2014 dataset has $9{,}998$ observations. 
Though the data sources are different, these two Guangzhou datasets both contain the rainy seasons of the city. We use the Beijing dataset as the target domain, which suffers from a large amount of ineffective training data due to the lack of rain. The Beijing dataset has $55{,}466$ observations for training, $3{,}000$ for validation, and $12{,}711$ for testing. All frames are resized to the resolution of $256\times256$.

\paragraph{Implementation.}
On all benchmarks, our final model has four stacked TMU layers with $64$-channel hidden states. We use the ADAM optimizer \cite{Kingma2014Adam} with a starting learning rate of $0.001$ for training the TMU network. Unless otherwise mentioned, the batch size is set to $8$, and the training process is stopped after $80{,}000$ iterations. All experiments are implemented in PyTorch \cite{PaszkeGMLBCKLGA19} and conducted on NVIDIA TITAN-RTX GPUs. We run all experiments three times and use the average results for quantitative evaluation. 
As for the dimensionality of the tensors, all the dimensions of the source and target states should be matched, including the number of channels ($P$), width ($W$), and height ($H$). For example, on the human motion benchmark, both $C_t^m$ (source, KTH/WEI) and $C_t$ (target, Human3.6M) are 3D tensors of $64\times 32\times 32$. We use a standard frame sub-scaling method \cite{shi2015convolutional} to transform the input images from $P\times W\times H$ to $(P\cdot K\cdot K) \times (W/K) \times (H/K)$ and control $K$ to make $W/K$ and $H/K$ constant across source and target domains.

\begin{table*}[t]
    \caption{Quantitative results on the flying digits benchmark. We use the $3$-digits subdataset as the target domain.
    A lower MSE or a higher SSIM per frame indicates better prediction results. All compared models are built upon the same ConvLSTM architecture.}
    \label{mnist_convlstm}
    \begin{center}
    \small
    \begin{sc}
    \renewcommand{\multirowsetup}{\centering}  
    \begin{tabular}{llccccccc}
    \toprule
    \multirow{2}{1cm}{Method} &
    \multirow{2}{1cm}{Sources} &
    \multirow{2}{1cm}{MSE} &
    \multirow{2}{1cm}{SSIM} &
    \multicolumn{2}{c}{\#Parameters} & \multicolumn{2}{c}{Runtime} \\
     & & & & Train & Test & Train & Test \\
    \midrule
    ConvLSTM \cite{shi2015convolutional} & None & 120.5 & 0.712 & 3.0M & 3.0M & 0.35s/batch & 16.3ms/seq \\
    TMU (train from scratch) & None & 120.6& 0.715 & - & - & - & - \\
    \midrule
    TMU (finetune) & 1 digit & 114.8 & 0.720 & 3.0M & 3.0M & 0.35s/batch & 16.3ms/seq \\
    TMU (finetune) & 2 digits & 110.0 & 0.732 & - & - & -& -\\
    L2SP \cite{LiGD18} & 1 digit & 118.5 & 0.703 & 3.0M & 3.0M & 0.39s/batch & 16.3ms/seq \\
    L2SP \cite{LiGD18} & 2 digits & 116.4 & 0.705 & - &- & -& -\\
    Knowledge Flow \cite{LiuPS19} & Both & 107.2 & 0.748 & 10.1M & 4.1M & 0.54s/batch & 21.7ms/seq \\
    ART \cite{CuiZSJW19} & Both & 105.0 & 0.734 & 10.1M & 10.1M & 0.73s/batch & 25.3ms/seq \\ 
    \midrule
    TMU (memory transfer) & 1 digit & 96.1 & 0.762 &-& - & - & -\\
    TMU (memory transfer) & 2 digits & 97.3 & 0.756 &-& - & - & -\\
    TMU (memory transfer) & Both & \textbf{94.7} & \textbf{0.777} & 9.5M & 3.6M & 0.43s/batch & 17.7ms/seq\\
    \bottomrule
    \end{tabular}
    \end{sc}
    \end{center}
\end{table*}

\begin{figure}[t]
    \vspace{-5pt}
    \centering
    \includegraphics[width=1\columnwidth]{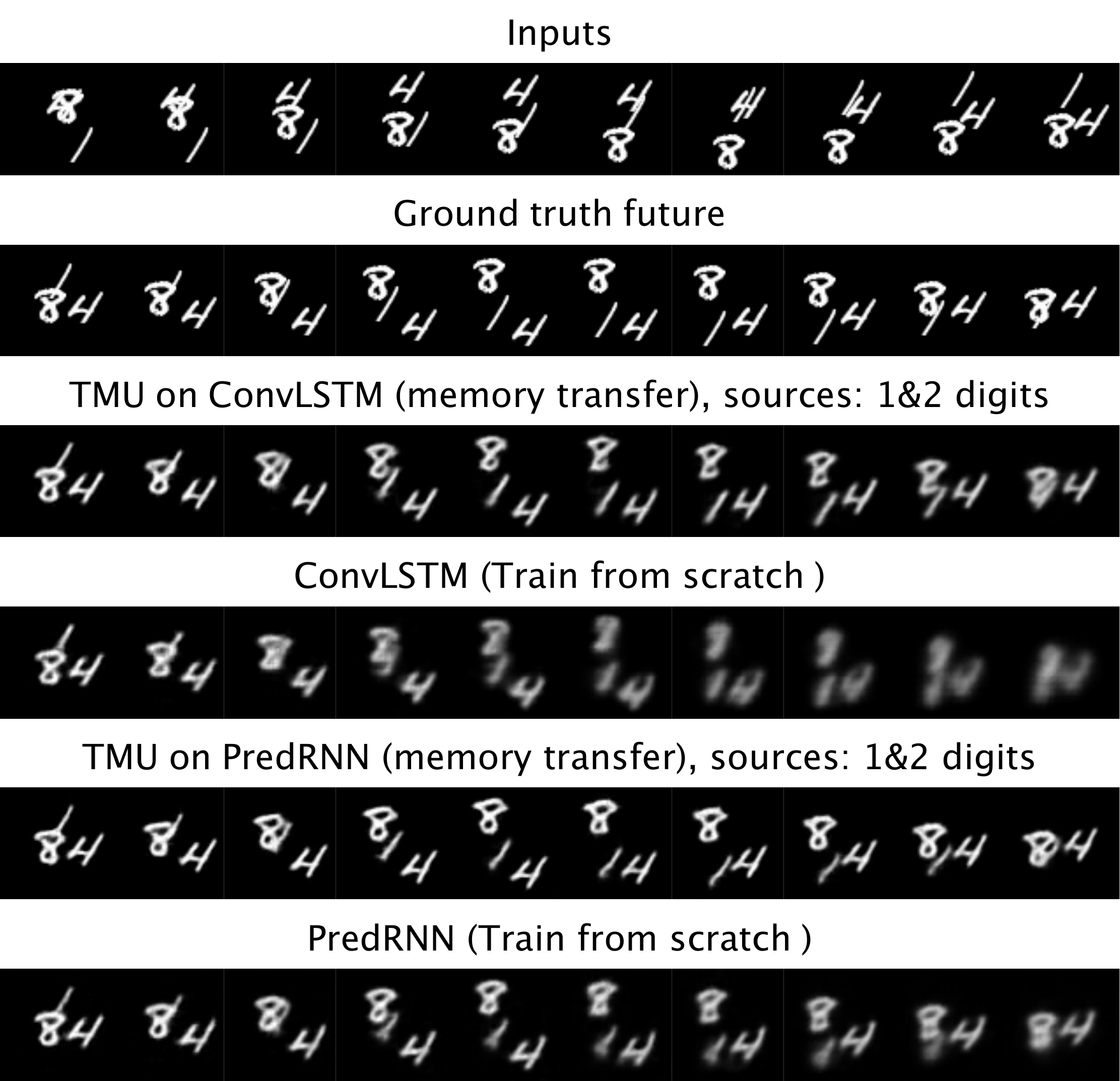}
    \vspace{-15pt}
    \caption{Predicted frames on the flying digits benchmark. Our transfer learning approach can consistently outperform the baseline predictive networks being trained from scratch.}
    \label{mnist_showcase}
\end{figure}

\subsection{Flying Digits Benchmark}

\paragraph{Setups.}
We take the $3$-digits Moving MNIST dataset as the target domain. Due to frequent occlusions and complex motions, it is challenging to accurately predict the trajectories of all three digits. We expect to improve this task by transferring the understandings of the digit's motion from the existing models that were pretrained with fewer digits. 

\paragraph{Comparing with training from scratch.}

\tab{mnist_convlstm} and \fig{mnist_showcase} respectively give the quantitative and qualitative results of our approach.
Compared with training a model on the target dataset from scratch, it gains significant improvements by learning from pretrained models on $1\&2$-digits Moving MNIST.
Besides, by comparing the training-from-scratch TMU model without any pretrained models with the training-from-scratch ConvLSTM network (the first two rows in \tab{mnist_convlstm}), we may conclude that it is the \emph{memory transfer} mechanism that improves the final results, instead of the engineering on the network architecture or the increased number of model parameters.

\paragraph{Comparing with previous transfer learning methods.}

Also shown in \tab{mnist_convlstm}, our approach outperforms finetuning by $17.5\%$ in MSE. It also achieves better results than existing transfer learning approaches, including L2SP \cite{LiGD18}, Knowledge Flow \cite{LiuPS19}, and ART \cite{CuiZSJW19}.
Furthermore, compared with finetuning, TMU with two sources only increases the number of parameters slightly at test time but improves MSE and SSIM remarkably. 
Compared with ART \cite{CuiZSJW19}, which is also particularly designed for RNNs, our approach only requires about one-third of the number of model parameters at test time. Thus, it does not increase the memory usage linearly with the growth of sources.
As for the training stage, all multi-source transfer learning models are forced to yield more parameters. A TITAN-RTX GPU can hold up to $41$ source ConvLSTM models and a target TMU model, which is sufficient for most practical application scenarios.

\paragraph{Backbones.} 
Our TMU can also be applied to other LSTM-based predictive models. We use PredRNN \cite{wang2017predrnn} and MIM \cite{WangZZLWY19} to take the place of the ConvLSTM network, covering both deterministic and stochastic models. Quantitative results and prediction examples are respectively shown in \tab{predrnn} and \fig{mnist_showcase}. 
The proposed TMU network achieves better results than directly finetuning the pretrained PredRNN or MIM on the $3$-digits dataset. It significantly improves the state-of-the-art MIM model in all metrics.

\begin{table}[t]
\caption{MSE/SSIM results of TMU upon different network backbones. We take the 3-digits Moving MNIST as the target domain.}
\label{predrnn}
\begin{center}
\small
\begin{sc}
\resizebox{\columnwidth}{!}{
\begin{tabular}{llcccc}
\toprule
Method & Sources & PredRNN & MIM \\
\midrule
From scratch & None & 93.4/0.802 & 89.0/0.783 \\
Finetune & 1 digit  & 91.1/0.811 & 84.5/0.794 \\
Finetune & 2 digits & 89.4/0.816 & 83.2/0.801 \\
Mem. transfer & Both & \textbf{84.9}/\textbf{0.828} & \textbf{75.3}/\textbf{0.838} \\
\bottomrule
\end{tabular}
}
\end{sc}
\end{center}
\end{table}

\begin{figure}[t]
    \centering
    \includegraphics[width=0.85\columnwidth]{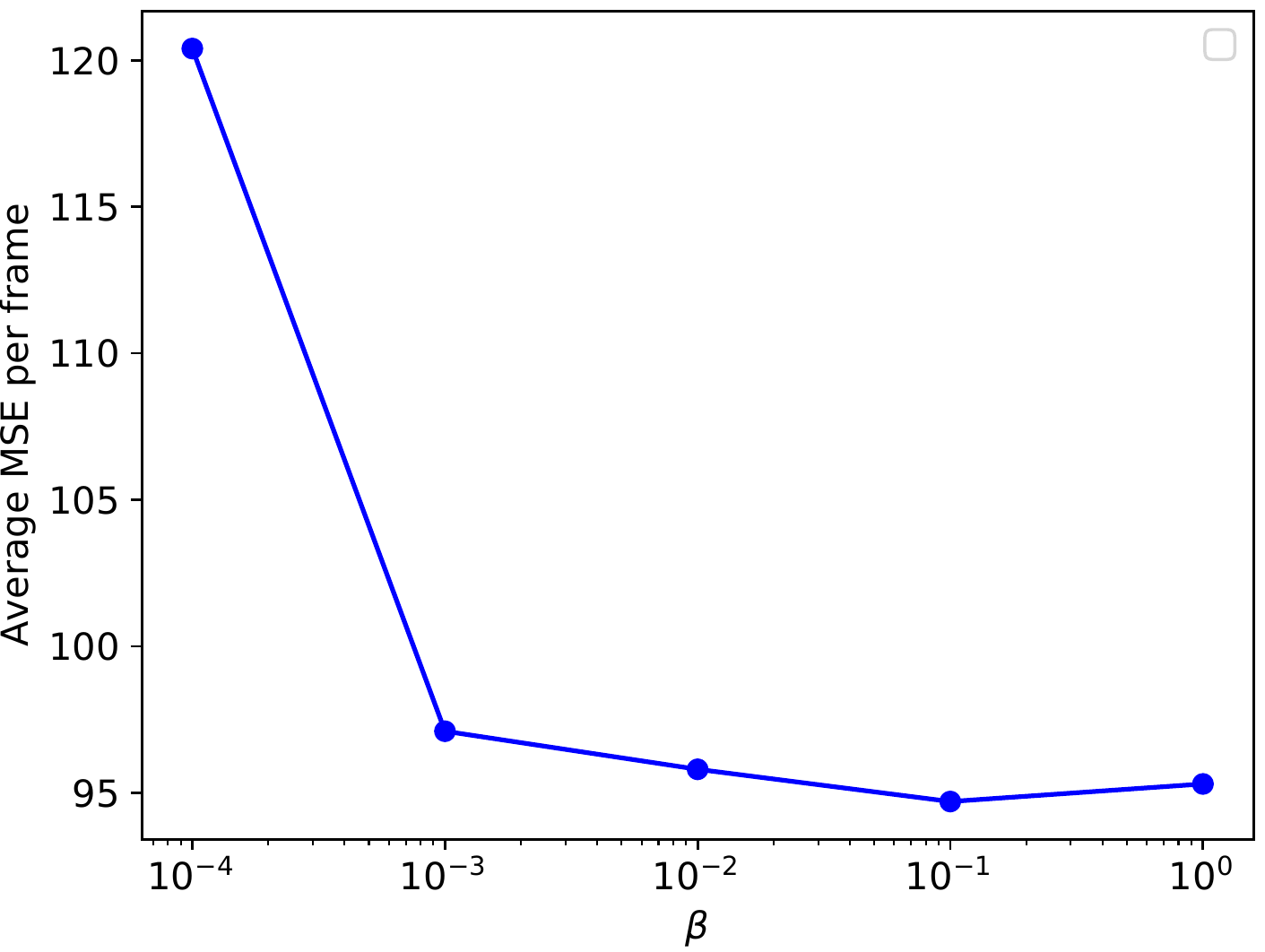}
    \vspace{-5pt}
    \caption{Sensitivity analysis of the hyper-parameter ($\beta$) for the unsupervised training objective on the flying digits benchmark.}
    \label{sensitivity}
    \vspace{-5pt}
\end{figure}

\paragraph{Hyper-parameters.} 
Last but not least, we show the sensitivity analysis of the training hyper-parameter $\beta$ in \fig{sensitivity}. It achieves the best results at $0.1$ on the flying digits benchmark and is robust and easy to tune in the range of $10^{-3}$ to $1$. We have similar results on the other two benchmarks and thus set $\beta$ to $0.1$ throughout this paper.

\begin{table*}[b]
    \caption{Quantitative results averaged per frame on Human3.6M using different network backbones, including ConvLSTM  \cite{shi2015convolutional}, MIM \cite{WangZZLWY19}, and SAVP \cite{lee2018stochastic}. For SAVP, we take the best one in SSIM from $100$ prediction samples.
    }
    \label{human_tab}
    \begin{center}
    \small
    \begin{sc}
    \renewcommand{\multirowsetup}{\centering}  
    \begin{tabular}{lllcccccc}
    \toprule
    \multirow{2}{1cm}{Model} &
    \multirow{2}{1cm}{Method} &
    \multirow{2}{1cm}{Sources} &
    \multicolumn{2}{c}{ConvLSTM} &
    \multicolumn{2}{c}{MIM} &
    \multicolumn{2}{c}{SAVP} 
    \\
    & & & MSE & SSIM & MSE & SSIM & MSE & SSIM \\
    \midrule
    \multirow{4}{1cm}{TMU} &
    Train from scratch & None & 504.2 & 0.762 & 430.5 & 0.790 & 465.2& 0.792 \\
    & Finetune & KTH & 472.0 & 0.778 & 420.1 & 0.796 & 453.7& 0.808\\
    & Finetune & Weizmann & 476.4 & 0.774 & 422.9 & 0.793 & 458.1& 0.805\\
    & Memory transfer & KTH \& Weizmann  & \textbf{442.5} & \textbf{0.794} & \textbf{394.2} & \textbf{0.813} & \textbf{430.2}& \textbf{0.831}\\
    \bottomrule
    \end{tabular}
    \end{sc}
    \end{center}
\end{table*}

\subsection{Human Motion Benchmark}

\paragraph{Setups.}

Compared with the KTH and Weizmann datasets with limited variability (they are small sets of backgrounds and actions, performed by a small group of individuals), the Human3.6M dataset contains larger amounts of data and more complex human motions, which makes it difficult to predict the future frames. 
On this benchmark, we take Human3.6M as the target domain and the other two datasets as the source domains.
We use ConvLSTM \cite{shi2015convolutional}, MIM \cite{WangZZLWY19}, and SAVP \cite{lee2018stochastic} as the network backbone of TMU, covering both deterministic and stochastic models.

\begin{figure}[t]
    \vspace{5pt}
    \centering
    \includegraphics[width=1\columnwidth]{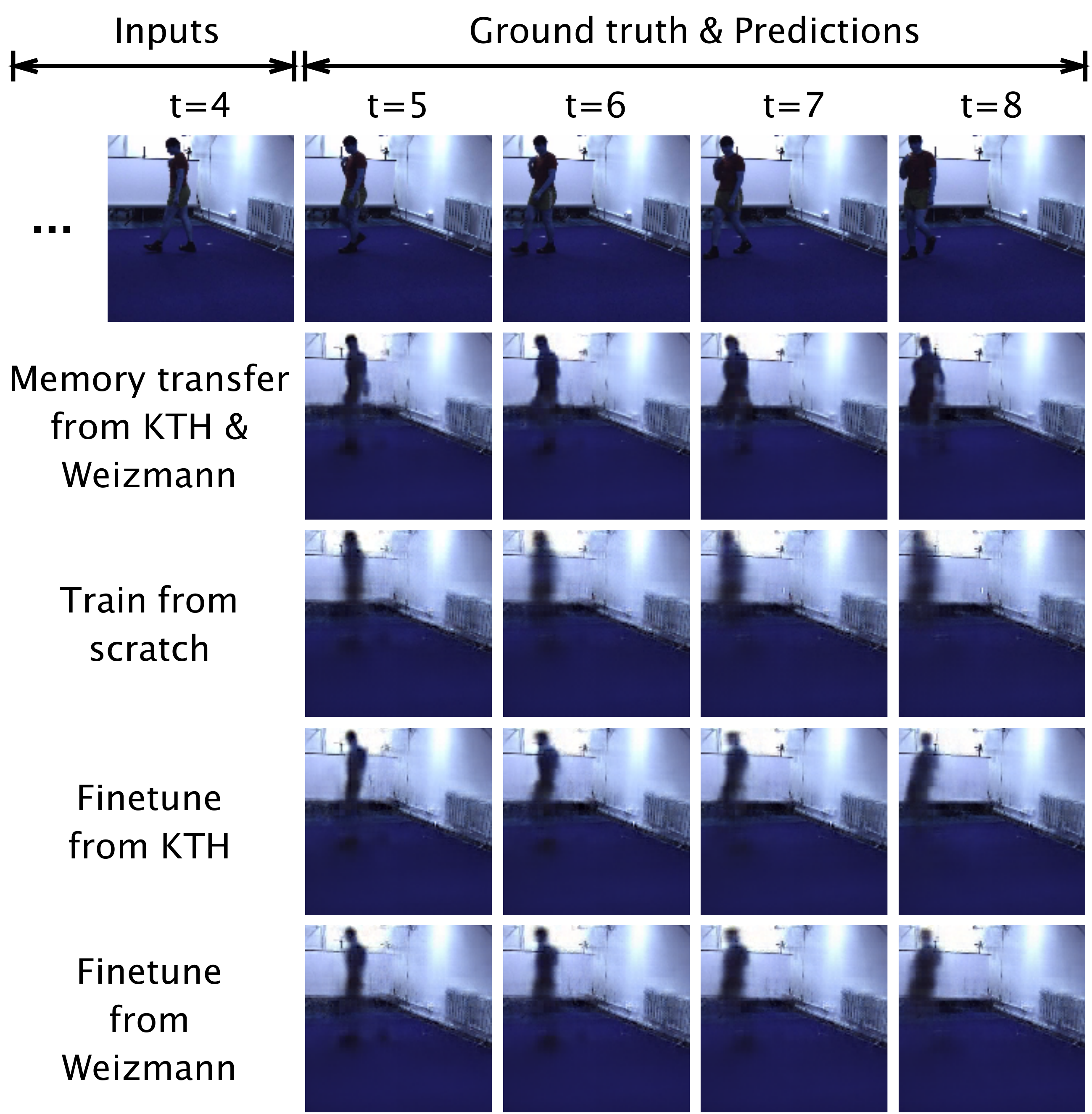}
    \vspace{-15pt}
    \caption{Prediction examples on Human3.6M by TMU networks based on ConvLSTM. Our method obtains the sharpest predictions.}
    \label{human_sample}
    \vspace{-5pt}
\end{figure}

\paragraph{Results.} 

We show the quantitative evaluations in \tab{human_tab}. The baseline TMU network, which takes either of the KTH or Weizmann datasets as the source domain, consistently outperforms the finetuning counterpart by large margins. 
The final TMU network that learns from both pretrained models further improves the prediction quality on the target task, which is because of the effectiveness of the \emph{transfer gates}. 
Besides, by using MIM and SAVP as the network backbones, we validate that TMU can outperform strong competitors that are pretrained well on the source datasets.
Moreover, we can see from \fig{human_sample} that the generated frames of the finetuning models largely suffer from blur effect, indicating that they are unable to capture a clear trend of motion. By contrast, the TMU network provides the sharpest results. We may conclude that the pretrained models from domains of plain backgrounds and simple actions can facilitate the training process of the model on a more challenging task, and the proposed \emph{transferable memory} framework can enhance this positive effect.

\subsection{Precipitation Nowcasting Benchmark}

\paragraph{Setups.}

\begin{figure}[t]
    \vspace{5pt}
    \centering
    \includegraphics[width=1\columnwidth]{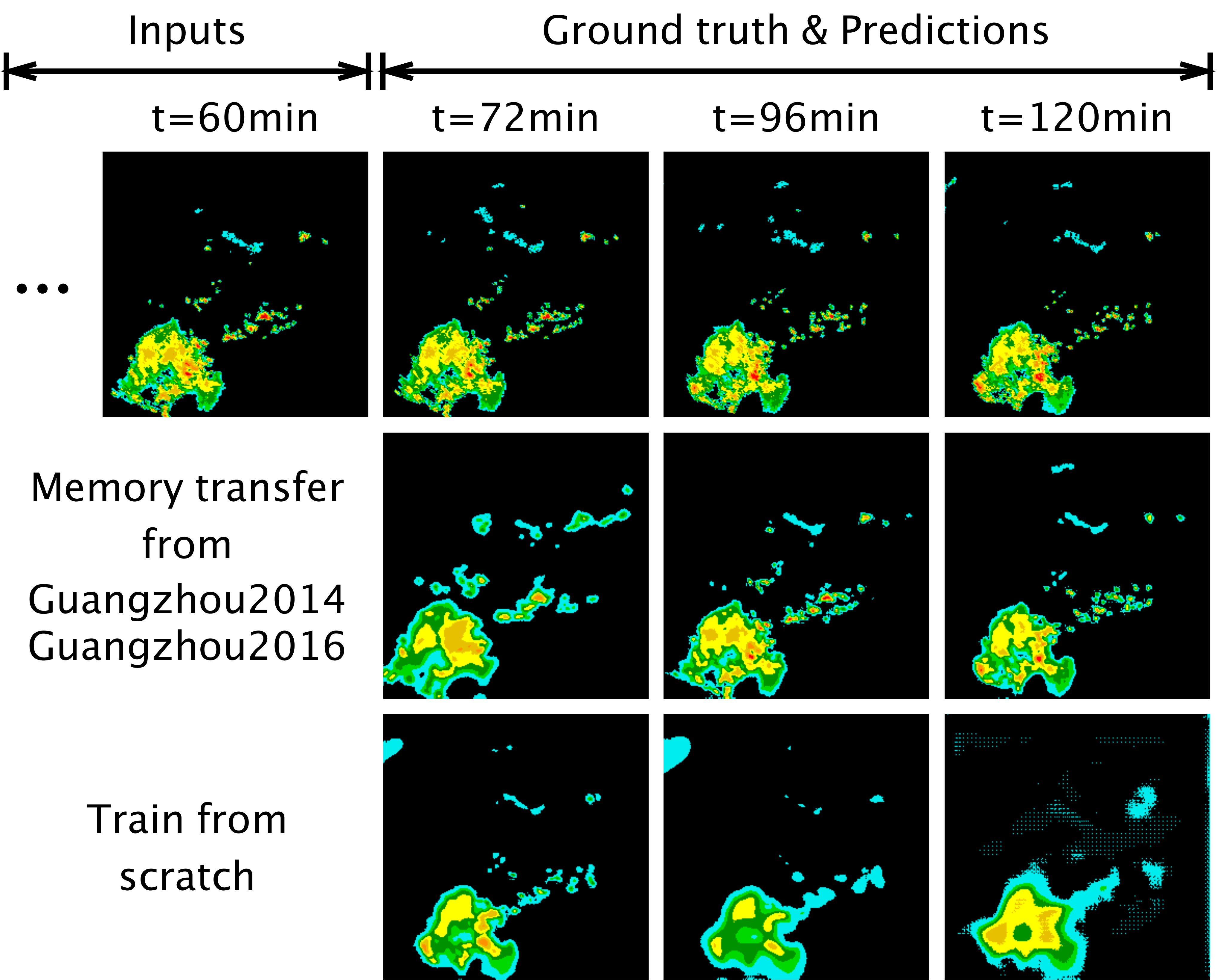}
    \vspace{-10pt}
    \caption{Prediction examples on the Beijing radar echo dataset by TMU networks based on ConvLSTM. We mainly compare the predicted high-intensity areas in yellow and red.}
    \label{radar_sample}
\end{figure}

We forecast the next $10$ radar echo frames from the previous $10$ observations, covering weather conditions in the next hour. Due to the lack of effective training data in the Beijing dataset, we take Guangzhou2014 and Guangzhou2016 as the source domains and pretrain models on these two datasets. Different from the previous experiments, the convolutional filters inside both the source ConvLSTM networks and the target TMU network are $3\times3$.

\paragraph{Results.}

In addition to MSE and MAE, here we also evaluate the predicted radar echoes using the Critical Success Index (CSI), which is defined as $\frac{\text { Hits }}{\text { Hits }+\text { Misses }+\text { FalseAlarms }}$. Here, hits correspond to the true positive, misses correspond to the false positive, and false alarms correspond to the false negative. We set the alarm threshold to $20$ dBZ. Compared with MSE and MAE, this metric is particularly sensitive to the high-intensity echoes. A higher CSI indicates better prediction results.
As shown in \tab{radar}, the TMU network remarkably outperforms the finetuning method in all evaluation metrics. \fig{radar_sample} provides showcases of predictions taking the Guangzhou radar echo datasets as source domains. Note that the ConvLSTM network without the help of the transferable memory framework makes fuzzy predictions, while the final TMU model forecasts the positions of high-intensity echoes (areas in red and yellow) more accurately.

\begin{table*}[t]
    \caption{Comparisons of transfer learning schemes within or across benchmarks using the Beijing radar echo dataset as the target domain.
    }
    \label{radar}
    \begin{center}
    \begin{small}
    \begin{sc}
    \renewcommand{\multirowsetup}{\centering}  
    \begin{tabular}{llcccc}
    \toprule
    Model & Method & Sources & MSE & MAE & CSI \\
    \midrule
    \multirow{8}{3cm}{TMU on ConvLSTM} &
    Train from scratch & None & 110.5 & 219.9 & 0.348 \\
    & Finetune & Guangzhou2014 & 96.6 & 198.3 & 0.368 \\
    & Finetune & Guangzhou2016 & 94.9 & 197.3 & 0.375 \\
    & Memory transfer & 2-digits MNIST \& KTH & 91.8 & 192.8 & 0.382 \\
    & Memory transfer & Guangzhou2014 & 87.4 & 193.3 & 0.374 \\
    & Memory transfer & Guangzhou2016 & 84.7 & 191.6 & 0.384 \\
    & Memory transfer & Guangzhou2014 \& 2016 & 77.3 & 184.2 & 0.403 \\
    & Memory transfer & Guangzhou \& MNIST \& KTH & \textbf{77.1} & \textbf{181.3} & \textbf{0.408} \\
    \bottomrule
    \end{tabular}
    \end{sc}
    \end{small}
    \end{center}
    \vspace{-5pt}
\end{table*}

\begin{table*}[t]
    \caption{The averages of transfer gate on the Beijing radar echo dataset. The model corresponds to the last one in \tab{radar}, where two relevant sources and two less relevant sources are used. The results show the significance of each source domain to the target domain.}
    \label{a_t}
    \vspace{5pt}
    \centering
    \small
    \begin{sc}
    \begin{tabular}{lcccc}
    \toprule
    Metric & Guangzhou2014 & Guangzhou2016 & 2-digits MNIST & KTH \\
    \midrule
    Values of Transfer gate ($a_t^m$) & 0.60 & 0.61 & 0.43 & 0.39 \\
    \bottomrule
    \end{tabular}
    \end{sc}
\end{table*}

\subsection{Further Analysis and Empirical Findings}
\label{findings}

\begin{figure}[t]
    \vspace{10pt}
    \centering
    \includegraphics[width=\columnwidth]{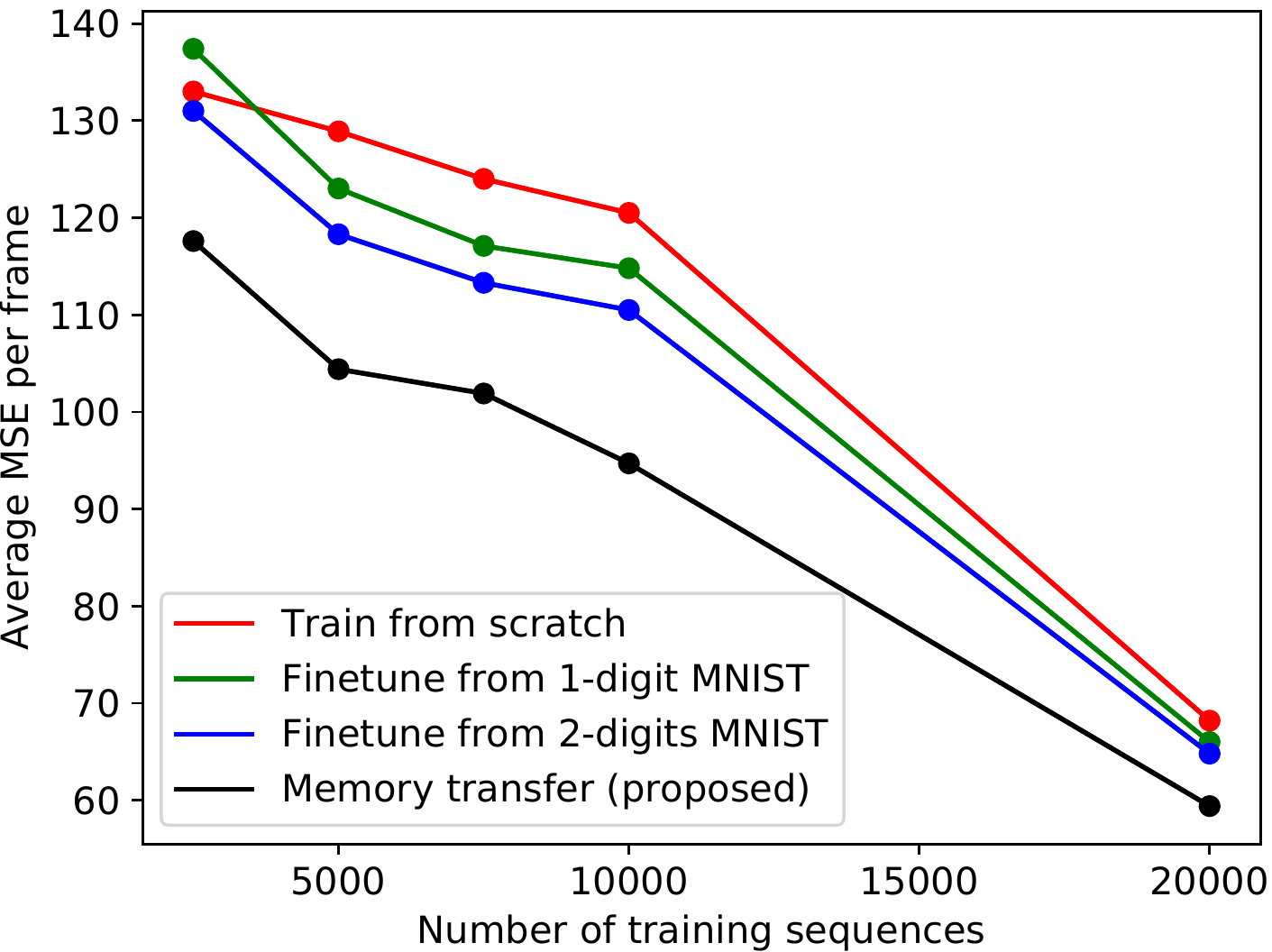}
    \vspace{-10pt}
    \caption{Averaged MSE per frame with respect to different numbers of training sequences of the target domain of the flying digits.
    }
    \label{sample_number}
\end{figure}

\paragraph{What if the target domain has sufficient training data?}

Based on the previous studies of transfer learning performed between supervised tasks, someone may concern that the effect of \emph{transferable memory} mechanism will degenerate when the number of training samples increases for the target domain. 
We explore this problem by training target models with respectively $25\%$, $50\%$, $75\%$, $100\%$, and $200\%$ training sequences for the $3$-digits Moving MNIST, and evaluate all the models on the same set (see \fig{sample_number}).
We observe the TMU network consistently outperforms the ones trained from scratch or finetuned upon the source models. 
Specifically, in the case that the target set has twice the training samples ($20{,}000$) than those in the standard settings, the finetuning method fails to remarkably improve the training-from-scratch baseline, while our approach achieves larger improvements. 
We may conclude that the main cause is that TMU enables successful distillation of the diverse understandings about the complex spatiotemporal dynamics, which can be a meaningful supplement to the target domain.

\begin{table}[t]
    \caption{Results of different transfer schemes using KTH as the target domain. All models are built upon the ConvLSTM network.}
    \label{kth_schemes}
    \vspace{5pt}
    \centering
    \small
    \begin{sc}
    \begin{tabular}{lccc}
    \toprule
    Method & Source & SSIM \\
    \midrule
    Train from scratch & None & 0.771 \\
    Memory transfer & Human3.6M  & 0.774 \\
    Memory transfer & 2-digits MNIST & 0.808 \\
    \bottomrule
    \end{tabular}
    \end{sc}
\end{table}

\paragraph{Finding favorable transfer schemes.}

We then explore the transfer schemes that can maximize the effectiveness of our method. What assumptions does our method make on the pretrained models?
\tab{kth_schemes} provides the results of different transfer schemes towards the KTH dataset. We observe that although the pretrained model on the KTH dataset can greatly help the training process on the Human3.6M dataset (SSIM: $0.762 \rightarrow 0.790$, \tab{human_tab}), conversely, the pretrained model from Human3.6M only has a slight effect on the KTH result (SSIM: $0.771 \rightarrow 0.774$, \tab{kth_schemes}). 
There are two possible causes: the first is that representations learned from the complex Human3.6M dataset do not have strong transferability; the second is that the training-from-scratch model on the KTH dataset is strong enough and cannot be further improved. 
To find the reason, we pretrain a model on the $2$-digits Moving MNIST dataset and apply it to the training process on KTH. We observe that such a transfer scheme obtains remarkable improvements over the baseline (SSIM: $0.771 \rightarrow 0.808$). Therefore, we can rule out the validity of the second hypothesis.
Since the Moving MNIST dataset only contains deterministic motions, the pretrained model yields less uncertainty about the future spatiotemporal dynamics. We may conclude that \emph{a favorable transfer scheme is to use the knowledge of better pretrained, more deterministic source models, and thus the final model can focus more on the domain-specific mode of target data.}

\paragraph{Will content-irrelevant source domains benefit the target predictive learning task?}

We take the radar echo dataset from the city of Beijing as the target domain, and the seemingly irrelevant $2$-digits Moving MNIST dataset and KTH dataset as the source domains. From \tab{radar}, we find that using KTH and Moving MNIST pretrained models can greatly help the prediction results, which outperforms the training-from-scratch TMU baseline (CSI: $0.348 \rightarrow 0.382$). 
Such results might be counter-intuitive, yet important to our understandings of the transferability of spatiotemporal modeling. 
Further, we take both the relevant Guangzhou2014 and Guangzhou2016 datasets as well as the seemingly irrelevant Moving MNIST dataset and KTH dataset as source domains. As opposed to our common sense for supervised transfer learning, TMU benefits from the less relevant sources, which are unrelated in image appearance but related in temporal dynamics (can be transferable). We then use the averages of $a_t^m$ to analyze the significance of each source domain. As shown in \tab{a_t}, TMU has higher $a_t^m$ for the Guangzhou radar echo datasets, indicating that it can adaptively control the influence of different sources via the transfer gates. 
Unlike supervised transfer learning, where irrelevant source data may cause negative effects, our approach can associate a variety of source domains and transfer temporal dynamics even if the content of source videos seems irrelevant.

\section{Conclusion and Discussion}

In this paper, we studied a new unsupervised transfer learning problem of using multiple pretrained models to improve the performance of a new spatiotemporal predictive learning task.
We used the term unsupervised for two reasons. First, we only explored the transfer learning cases between multiple unsupervised tasks. Second, the proposed method does not require any labels.
We proposed the transferable memory framework, which transfers knowledge from multi-source RNNs and yielded better results than finetuning. Our approach was shown effective even in the case that there is adequate data for the target domain, or the pretrained models were collected from less relevant domains.
Code and datasets are made available at \url{https://github.com/thuml/transferable-memory}.

One potential work in the future is to explore how to transfer knowledge between unsupervised tasks beyond predictive learning. Another one is that how we can transfer knowledge from unsupervisedly learned RNN models to CNN models of the downstream supervised task.
Our approach is also likely to be effective for supervised tasks, though it may not be the best choice when labels are available. It is worth exploring, but beyond the scope of the paper.

\section*{Acknowledgements}
This work was supported by the Natural Science Foundation of China ($61772299$, $71690231$), and China University S\&T Innovation Plan Guided by the Ministry of Education.

\bibliography{icml2020}
\bibliographystyle{icml2020}

\end{document}